\documentclass[final]{cvpr}

\usepackage{times}
\usepackage{epsfig}
\usepackage{graphicx}
\usepackage{amsmath}
\usepackage{amssymb}

\usepackage{verbatim}
\usepackage[ruled,linesnumbered]{algorithm2e}
\usepackage{booktabs}
\usepackage[flushleft]{threeparttable}
\usepackage[titletoc]{appendix}
\usepackage{etoolbox}
\usepackage[pagebackref=true,breaklinks=true,colorlinks,bookmarks=false]{hyperref}

\def\cvprPaperID{2427} % *** Enter the CVPR Paper ID here
\def\confYear{CVPR 2021}

\begin{document}

%%%%%%%%% TITLE
\title{Learning a Proposal Classifier for Multiple Object Tracking}  % **** Enter the paper title here

\author{Peng Dai$^{1}$ \and Renliang Weng$^{2}$ \and Wongun Choi$^{2}$ \and Changshui Zhang$^{1}$ \and Zhangping He$^{2}$ \and  Wei Ding$^{2}$ \\
$^{1}$Tsinghua University, Beijng, China. \hspace{0.6cm} $^{2}$Aibee Inc \\
{\tt\small $^{1}$\{daip2020, zcs\}@mail.tsinghua.edu.cn \hspace{1cm} $^{2}$\{rlweng, wgchoi, zphe, weiding\}@aibee.com}
}

\maketitle

\begin{figure*}[htb]
\centering
\includegraphics[width=0.90\textwidth]{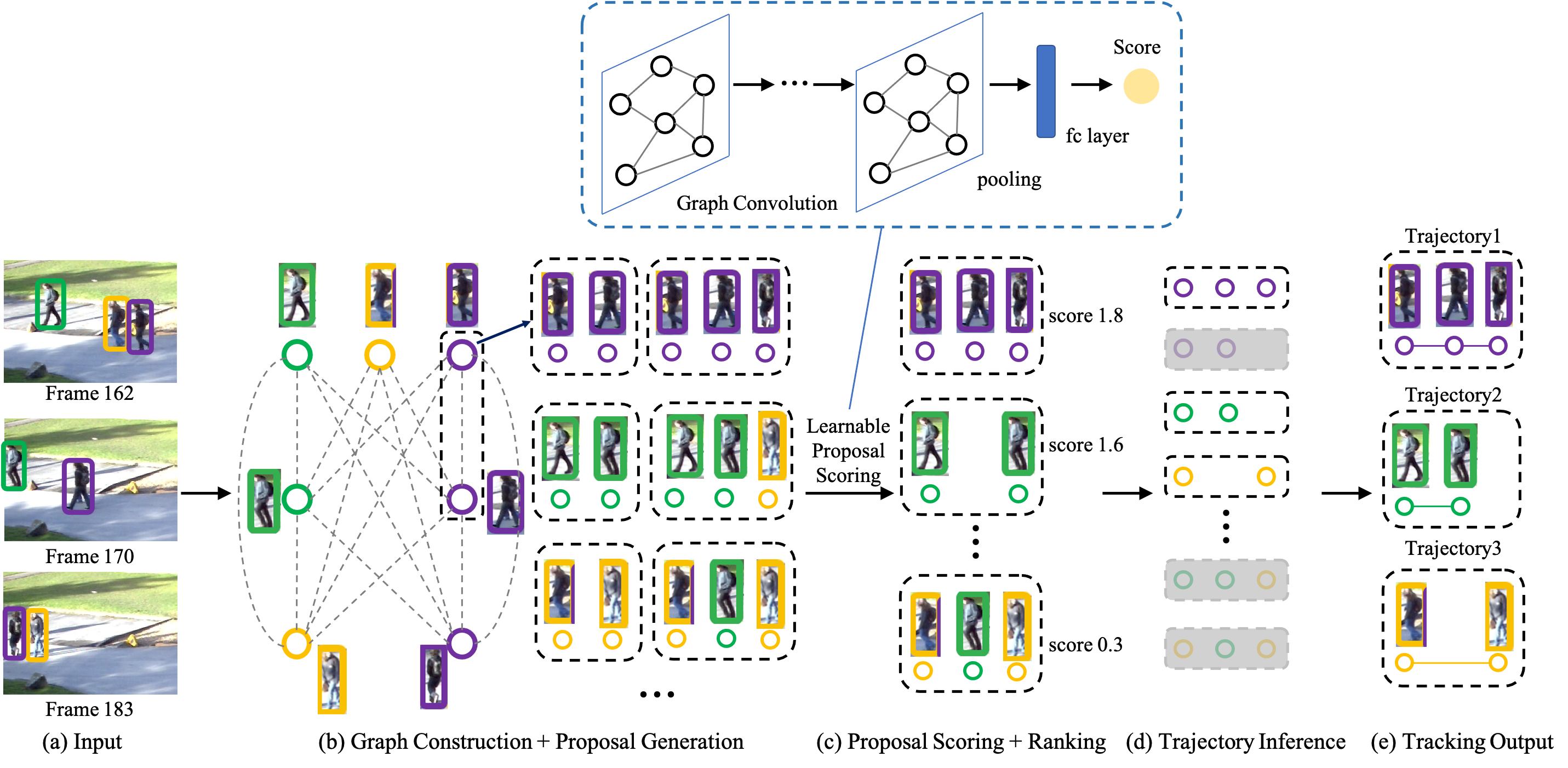}
\caption{Overview of our framework. (a) Given a set of frames and detections as input. (b) A graph is constructed to model the data association problem. Nodes in the graph represent detections/tracklets and the edges indicate possible links among nodes. The nodes in different colors represent different objects. Similar to two-stage object detector faster RCNN, our method adopts a proposal-based framework.
Multiple proposals (i.e., candidate object trajectories) are generated based on the affinity graph. (c) We evaluate the quality scores for the generated proposals with trainable GCN. (d) A simple de-overlapping strategy is adopted to do trajectory inference and (e) obtain the final tracking output.}
\label{fig:picture000}
\end{figure*}

%%%%%%%%% ABSTRACT
\begin{abstract}
   The recent trend in multiple object tracking (MOT) is heading towards leveraging deep learning to boost the tracking performance. However, it is not trivial to solve the data-association problem in an end-to-end fashion. In this paper, we propose a novel proposal-based learnable framework, which models MOT as a proposal generation, proposal scoring and trajectory inference paradigm on an affinity graph. This framework is similar to the two-stage object detector Faster RCNN, and can solve the MOT problem in a data-driven way. For proposal generation, we propose an iterative graph clustering method to reduce the computational cost while maintaining the quality of the generated proposals. For proposal scoring, we deploy a trainable graph-convolutional-network (GCN) to learn the structural patterns of the generated proposals and rank them according to the estimated quality scores. For trajectory inference, a simple deoverlapping strategy is adopted to generate tracking output while complying with the constraints that no detection can be assigned to more than one track. We experimentally demonstrate that the proposed method achieves a clear performance improvement in both MOTA and IDF1 with respect to previous state-of-the-art on two public benchmarks. 
   %Our code will be released soon.
   Our code is available at \url{https://github.com/daip13/LPC_MOT.git}.
\end{abstract}

%%%%%%%%% BODY TEXT - ENTER YOUR RESPONSE BELOW
\section{Introduction} \label{section:introduction}

Tracking multiple objects in videos is an important problem in many application domains.
Particularly, estimating humans location and their motion is of great interest in surveillance, business analytics, robotics and autonomous driving.
Accurate and automated perception of their whereabouts and interactions with others or environment can help identifying potential illegal activities, understanding customer interactions with retail spaces, planning the pathway of robots or autonomous vehicles.

The ultimate goal of multiple object tracking (MOT) is to estimate the trajectory of each individual person as one complete trajectory over their whole presence in the scene without having any contamination by the others. Much research is done in this domain to design and implement robust and accurate MOT algorithms in the past~\cite{braso2020learning, kim2015multiple, schulter2017deep}. However, the problem still remains unsolved as reported in the latest results in various public benchmarks~\cite{dave2020tao, dendorfer2020mot20, geiger2013vision, milan2016mot16}. 
The key challenges in MOT are mostly due to occlusion and scene clutter, as in any computer vision problem. Consider the case when two people (yellow and purple boxes in Fig.~\ref{fig:picture000}) are walking together in a spatial neighborhood. At one point, both people are visible to the camera and recent object detection algorithms like~\cite{lin2017focal, redmon2018yolov3, ren2015faster}, can easily detect them. When the two people become aligned along the camera axis, however, one is fully occluded by another, and later both become visible when one passes the other. Since the visual appearance may have subtle difference between the two targets due to various reasons like illumination, shading, similar clothing, etc, estimating the trajectory accurately without contamination (often called as identity transfer) remains as the key challenge. In more crowded scenes, such occlusion can happen across multiple peoples which pose significant troubles to any MOT algorithm. Moreover, the MOT problem naturally has an exponentially large search space for the solution~\footnote{The tracking-by-detection approach, which is the de-facto framework in MOT domain, needs to solve the data-association problem given detections at each timestamp. The size of hypothesis space is exponential to the number of detections~\cite{kim2015multiple}.} which prohibits us from using complicated mechanisms.

Traditional approaches focus on solving the problem by employing various heuristics, hand-defined mechanisms to handle occlusions~\cite{brendel2011multiobject, kim2015multiple}.
Multiple Hypotheses Tracking (MHT~\cite{kim2015multiple}) is one of the earliest successful algorithms for MOT.
A key strategy in MHT to handle occlusions is to delay data-association decisions by keeping multiple hypotheses active until data-association ambiguities are resolved.
Network flow-based methods~\cite{brendel2011multiobject, butt2013multi} have recently become a standard approach for MOT due to their computational efficiency and optimality.
In this framework, the data-association problem is modeled as a graph, where each node represents a detection and each edge indicates a possible link between nodes.
Then, occlusions can be handled by connecting non-consecutive node pairs.
Both MHT and network flow-based methods need to manually design appropriate gap-spanning affinity for different scenarios.
However, it is infeasible to enumerate all possible challenging cases and to implement deterministic logic for each case.
%Recently, several works are proposed to employ learning based mechanism to solve the MTT problem~\cite{braso2020learning,xu2020train,schulter2017deep, sadeghian2017tracking}. Schulter et al.~\cite{schulter2017deep} proposed a bi-level optimization framework to learn features for network-flow-based data association in an end-to-end fashion. Xu et al.~\cite{xu2020train} proposed a differentiable Deep Hungarian Net (DHN) to approximate the Hungarian (Munkres) algorithm and provide a soft approximation of the optimal prediction-to-ground-truth assignment. But all these works are limited to learning pairwise costs, and cannot incorporate higher-order information across nodes. To tackle this problem, Bras\'o et al.~\cite{braso2020learning} proposed a fully differentiable framework based on Message Passing Networks (MPNs), which can exploit the natural graph structure of the problem to perform both feature learning and final solution prediction. However, this pipeline does not generally guarantee the \textit{flow conservation constraints}\cite{ahyja1993network}. An extra greedy rounding scheme is needed to obtain a feasible binary output.

In this paper, we propose a simple but surprisingly effective method to solve the MOT problem in a data-driven way. Inspired by the latest advancement in object detection~\cite{ren2015faster} and face clustering~\cite{yang2019learning}, we propose to design the MOT algorithm using two key modules, 1) proposal generation and 2) proposal scoring with graph convolutional network (GCN)~\cite{kipf2016semisupervised}. Given a set of short tracklets (locally grouped set of detections using simple mechanisms), our proposal generation module (see Fig.~\ref{fig:picture000}(b)) generates a set of proposals that contains the complete set of tracklets for fully covering each individual person, yet may as well have multiple proposals with contaminated set of tracklets (i.e., multiple different people merged into a proposal). The next step is to identify which proposal is better than the others by using a trainable GCN and rank them using the learned ranking/scoring function (see Fig.~\ref{fig:picture000}(c)). Finally, we adopt an inference algorithm to generate tracking output given the rank of each proposal (see Fig.~\ref{fig:picture000}(d)), while  complying with the typical tracking constraints like no detection assigned to more than one track. 
%Our experimental evaluation on two challenging public benchmarks shows a significant improvement in both MOTA and IDF1 measures.

The main contribution of the paper is in four folds: 
1) We propose a novel learnable framework which formulates MOT as a proposal generation, proposal scoring and trajectory inference pipeline. In this pipeline, we can utilize algorithms off the shelf for each module. 
2) We propose an iterative graph clustering strategy for proposal generation. It can significantly reduce the computational cost while guaranteeing the quality of the generated proposals.
3) We employ a trainable GCN for proposal scoring.
By directly optimizing the whole proposal score rather than the pairwise matching cost, GCN can incorporate higher-order information within the proposal to make more accurate predictions.
4) We show significantly improved state-of-the-art results of our method on two MOTChallenge benchmarks.

%The paper is organized as follows:
%Section \ref{section:related_works} gives a brief review of the related works.
%Section \ref{section:method} describes details of the proposed algorithm.
%Section \ref{section:experiment} reports a comprehensive set of comparison experiments. Finally, Section \ref{section:conclusion} summarizes this paper.

\section{Related Work} \label{section:related_works}
Most state-of-the-art MOT works follow the tracking-by-detection paradigm which divides the MOT task into two sub-tasks: first, obtaining frame-by-frame object detections; second, linking the set of detections into trajectories.
The first sub-task is usually addressed with object detectors~\cite{lin2017focal, redmon2018yolov3, ren2015faster, yang2016exploit}.
While the latter can be done on a frame-by-frame basis for online applications~\cite{hu2020multi, Wojke2017simple, xu2019spatial, zhou2018deep, zhu2018online} or a batch basis for offline scenarios~\cite{berclaz2011multiple, braso2020learning, milan2015multi}.
For video analysis tasks that can be done offline, batch methods are preferred since they can incorporate both past and future frames to perform more accurate association and are more robust to occlusions.
A common approach to model data-association in a batch manner is using a graph, where each node represents a detection and each edge indicates a possible link between nodes.
Then, data-association can be converted to a graph partitioning task, i.e., finding the best set of active edges to predict partitions of the graph into trajectories.
Specifically, batch methods differ in the specific optimization methods used, including network flow \cite{Pirsiavash2011Globally}, generalized maximum multi clique \cite{dehghan2015gmmcp}, linear programming \cite{jiang2007linear}, maximum-weight independent set \cite{brendel2011multiobject}, conditional random field \cite{yang2012online}, k-shortest path \cite{berclaz2011multiple}, hyper-graph based optimization \cite{Wen2014Multiple}, etc.
However, the authors in \cite{bergmann2019tracking} showed that the significantly higher computational cost of these overcomplicated optimization methods does not translate to significantly higher accuracy.

%\textbf{Learning in MOT.}
As summarized in \cite{ciaparrone2020deep, leal2017tracking}, the research trend in MOT has been shifting from trying to find better optimization algorithms for the association problem to focusing on the use of deep learning in affinity computation.
Most existing deep learning MOT methods focus on improving the affinity models, since deep neural networks are able to learn powerful visual and kinematic features for distinguishing the tracked objects from the background and other similar objects.
Leal{-}Taix{\'{e}} et al.~\cite{leal2016learning} adopted a Siamese convolutional neural network (CNN) to learn appearance features from both RGB images and optical flow maps.
Amir et al.~\cite{sadeghian2017tracking} employed long short-term memory (LSTM) to encode long-term dependencies in the sequence of observations.
Zhu et al.~\cite{zhu2018online} proposed dual matching attention networks with both spatial and temporal attention mechanisms to improve tracking performance especially in terms of identity-preserving metrics.
Xu et al.~\cite{xu2019spatial} applied spatial-temporal relation networks to combine various cues such as appearance, location, and topology.
Recently, the authors in \cite{bergmann2019tracking, ristani2018features} confirmed the importance of learned re-identification (ReID) features for MOT.
All aforementioned methods learn the pair-wise affinities independently from the association process, thus a classical optimization solver is still needed to obtain the final trajectories.

Recently, some works~\cite{braso2020learning, chu2019famnet, schulter2017deep, xu2020train} incorporate the optimization solvers into learning.
Chu et al.~\cite{chu2019famnet} proposed an end-to-end model, named FAMNet, to refine feature representation, affinity model and multi-dimensional assignment in a single deep network.
Xu et al.~\cite{xu2020train} presented a differentiable Deep Hungarian Net (DHN) to approximate the Hungarian matching algorithm and provide a soft approximation of the optimal prediction-to-ground-truth assignment.
%Xu et al.~\cite{xu2020train} presented a Deep Hungarian Net (DHN) module to approximate the Hungarian matching algorithm, and two differentiable proxies of MOTA and MOTP to optimize the deep tracker.
Schulter et al.~\cite{schulter2017deep} designed a bi-level optimization framework which frames the optimization of a smoothed network flow problem as a differentiable function of the pairwise association costs.
%Computation of the optimal association contains non-differentiable operations. To make it differentiable, these approaches formulate the association problem as pairwise cost minimization problem.
Bras\'o et al.~\cite{braso2020learning} modeled the non-learnable data-association problem as a differentiable edge classification task.
In this framework, an undirected graph is adopted to model the data-association problem.
Then, feature learning is performed in the graph domain with a message passing network.
Next, an edge classifier is learned to classify edges in the graph into active and non-active.
%An active edge indicates the linked two detections belong to the same trajectory.
Finally, the tracking output is efficiently obtained via grouping connected components in the graph.
However, this pipeline does not generally guarantee the \textit{flow conservation constraints}~\cite{ahyja1993network}.
The final tracking performance might be sensitive to the percentage of flow conservation constraints that are satisfied.

Similar to \cite{braso2020learning}, our method also models the data-association problem with an undirected graph.
However, our approach follows a novel proposal-based learnable MOT framework, which is similar to the two-stage object detector Faster RCNN~\cite{ren2015faster}, i.e. proposal generation, proposal scoring and proposal pruning. 
%Specifically, our framework is different from \cite{braso2020learning} in the following two aspects. First, the \textit{flow conservation constraints} are guaranteed in the proposal generation step. Second, by optimizing the whole proposal score rather than the pairwise matching cost, the learned GCN can take higher-order information within the proposal into consideration to make globally informed predictions.

\section{Method} \label{section:method}

Given a batch of video frames and corresponding detections $\mathcal{D}=\{d_{1}, \cdots ,d_{k}\}$, where $k$ is the total number of detections for all frames. Each detection is represented by $d_{i} = (o_{i}, p_{i}, t_{i})$, where $o_{i}$ denotes the raw pixels of the bounding box, $p_{i}$ contains its 2D image coordinates and $t_{i}$ indicates its timestamp.
A trajectory is defined as a set of time-ordered detections $\mathcal{T}_{i} = \{d_{i_{1}}, \cdots ,d_{i_{n_{i}}}\}$, where $n_{i}$ is the number of detections that form trajectory $i$.
The goal of MOT is to assign a track ID to each detection, and form a set of $m$ trajectories $\mathcal{T}_{*}=\{\mathcal{T}_{1}, \cdots ,\mathcal{T}_{m}\}$ that best maintains the objects' identities.

\subsection{Framework Overview}\label{subsectionFO}
%In this paper, we propose a novel learnable framework for MOT, which adopts a ``proposal generation-proposal scoring-inference'' pipeline.
As shown in Figure~\ref{fig:picture000}, our framework consists of four main stages.
%This framework is similar to the two-stage object detector Faster RCNN~\cite{ren2015faster}, and consists of four main stages.

\textbf{Data Pre-Processing.} 
To reduce the ambiguity and computational complexity in proposal generation, a set of tracklets $\mathcal{T}=\{\mathcal{T}_{1}, \cdots ,\mathcal{T}_{n}\}$ is generated by linking detections $\mathcal{D}$ in consecutive frames.
And these tracklets $\mathcal{T}$ are utilized as basic units in downstream modules. 

\textbf{Proposal Generation.}
As shown in Figure~\ref{fig:picture000}(b), we adopt a graph $\mathcal{G} = (\mathcal{V},\mathcal{E})$, where $\mathcal{V} :=\{v_{1}, \cdots, {v}_{n} \}$, $\mathcal{E} \subset \mathcal{V} \times \mathcal{V}$, to represent the structured tracking data $\mathcal{T}$. 
A proposal $\mathcal{P}_{i}$ = $\{ {v}_{i} \}$ is a subset of the graph $\mathcal{G}$.
% The objective of proposal generation is to obtain the perfect proposal $\hat{\mathcal{P}_{i}}$ which contains the complete set of track-lets for each object.
The objective of proposal generation is to obtain an over-complete set of proposals which contain at least one perfect proposal for each target.
However, it is computationally prohibitive to explore all perfect proposals $\{ \hat{\mathcal{P}_{i}} \}_{i=1}^{m}$ from the affinity graph $\mathcal{G}$.
Inspired by \cite{yang2019learning}, we propose an iterative graph clustering strategy in this paper.
By simulating the bottom-up clustering process, it can provide a good trade-off between proposal quality and the computational cost.

\textbf{Proposal Scoring.}
With the over-complete set of proposals $\mathcal{P} = \{ \mathcal{P}_{i} \}$, we need to calculate their quality scores and rank them, in order to select the subset of proposals that best represent real tracks.
Ideally, the quality score can be defined as a combination of precision and recall rates.
\begin{align}
    score(\mathcal{P}_{i}) &= rec(\mathcal{P}_{i}) + w \cdot prec(\mathcal{P}_{i}) \label{equa:quality_score} \\
    rec(\mathcal{P}_{i}) &= \frac{|\mathcal{P}_{i} \cap \hat{\mathcal{P}_{i}}|}{|\hat{\mathcal{P}_{i}}|} \label{equa:metriccomple} \\
    prec(\mathcal{P}_{i}) &= \begin{cases}
   1,  &if \ \  n( \mathcal{P}_{i} ) = 1\\
   0,  &otherwise  \label{equa:metricpurity}
   \end{cases}
\end{align}
where $w$ is a weighting parameter controlling the contribution of precision score, $\hat{\mathcal{P}_{i}}$ is the ground-truth set of all detections with label $major( \mathcal{P}_{i} )$, and $major( \mathcal{P}_{i} )$ is the majority label of the proposal $\mathcal{P}_{i}$, $\left | \cdot \right |$ measures the number of detections, $n( \mathcal{P}_{i} )$ represents the number of labels included in proposal $\mathcal{P}_{i}$.
Intuitively, $prec$ measures the purity, and $rec$ reflects how close $\mathcal{P}_{i}$ is to the matched ground-truth $\hat{\mathcal{P}_{i}}$.
Inspired by~\cite{yang2019learning}, we adopt a GCN based network to learn to estimate the proposal score given the above definition. %(Section~\ref{subsectionPCN}) 
The precision of a proposal can be learned with a binary-cross-entropy loss through training procedure.
\begin{comment}
\begin{equation}\label{equa:CrossEntropyLoss}
    L = \frac{1}{N}\sum_{i=1}^{N}-[l( \mathcal{P}_{i} ) log(p( \mathcal{P}_{i} ) + (1-l( \mathcal{P}_{i} ))(1 - log(p( \mathcal{P}_{i} ))]
\end{equation}
where $l( \mathcal{P}_{i})$ is the GT purity label of proposal $\mathcal{P}_{i}$, and $p( \mathcal{P}_{i} )$ denotes the probability that the GCN model predicts $\mathcal{P}_{i}$ to be pure.
\end{comment}
However, it is much harder for a GCN to learn the recall of a proposal without exploring the entire graph structure including the vertices that are very far from a given proposal. We find that the normalized track length ($\left | \mathcal{P}_{i} \right |/C$, where $C$ is a constant for normalization) is positively correlated with the recall of a proposal when precision is high. Thus, we approximate the recall rate of a proposal with the normalized track length and let the network to focus on accurately learning the precision of a proposal. 
%We fix the scalar weight $w=100$ throughout all the experiments.

\textbf{Trajectory Inference:}
Similar to the Non-Maximum Suppression in object detection, a trajectory inference strategy is needed to generate the final tracking output $\mathcal{T}_{*}$ with the ranked proposals.
This step is to comply with the tracking constraints like no tracklet assigned to more than one track.
To reduce the computational cost, we adopt a simple de-overlapping algorithm with a complexity of $O$($n$).

\subsection{Data Pre-processing}\label{subsectionDP}
A tracklet is widely used as an intermediate input in many previous works~\cite{dai2018instance,yoo2016online}. In our framework, we also use tracklets $\mathcal{T}=\{\mathcal{T}_{1}, \cdots ,\mathcal{T}_{n}\}$ as basic units for graph construction, where $n$ is the number of tracklets and is far less than detections $k$.
Hence, it can significantly reduce overall computation. 
%In this subsection, we provide a brief introduction to generate track-lets $\mathcal{T}$ from the input detections $\mathcal{D}$.
First, the ReID features $a_{i}$ for each detection $d_{i}$ is extracted with a CNN.
Then, the overall affinity of two detections or detection-to-tracklet is computed by accumulating three elementary affinities based on their appearance, timestamps and positions.
Finally, low-level tracklets are generated by linking detections based on their affinities with Hungarian algorithm~\cite{munkres1957algorithms}.
It is worth noting that the purity of the generated tracklets is crucial, because the downstream modules use them as basic units and there is no strategy to recover from impure tracklets. Similarly to \cite{huang2008robust}, we use a dual-threshold strategy in which a higher threshold $\theta_{1} $ is used to accept only associations with high affinities, and a lower threshold $\theta_{2} $ is to avoid associations that have rivals with comparable affinities. % With this dual-threshold method, majority of the the generated track-lets are pure.

\subsection{Iterative Proposal Generation}\label{subsectionPG}
\begin{figure}
\centering
\includegraphics[width=0.4\textwidth]{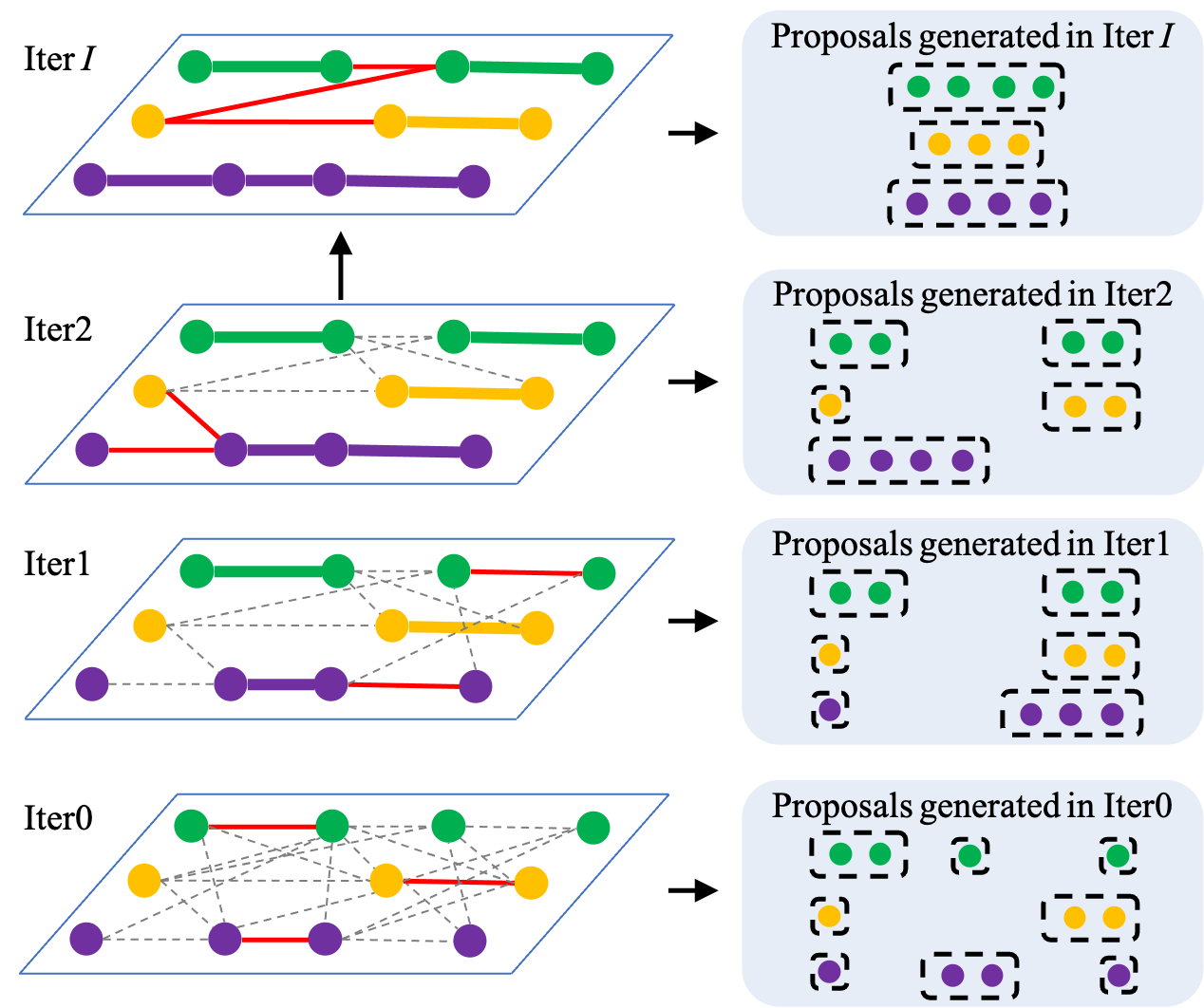}
\caption{Visualization of the iterative proposal generation. In each iteration, only a small part of edges (red solid line) that meet the gating thresholds can be active. Each cluster generated in iteration $i$ will be grouped as a vertex in iteration $i+1$. To keep the purity of the clusters, strict gating thresholds are set in the first few iterations. As iterations increase, these thresholds will be gradually relaxed to grow proposals.}
\label{fig:picture001}
\end{figure}

\begin{figure*}[tb]
\centering
\includegraphics[width=0.9\textwidth]{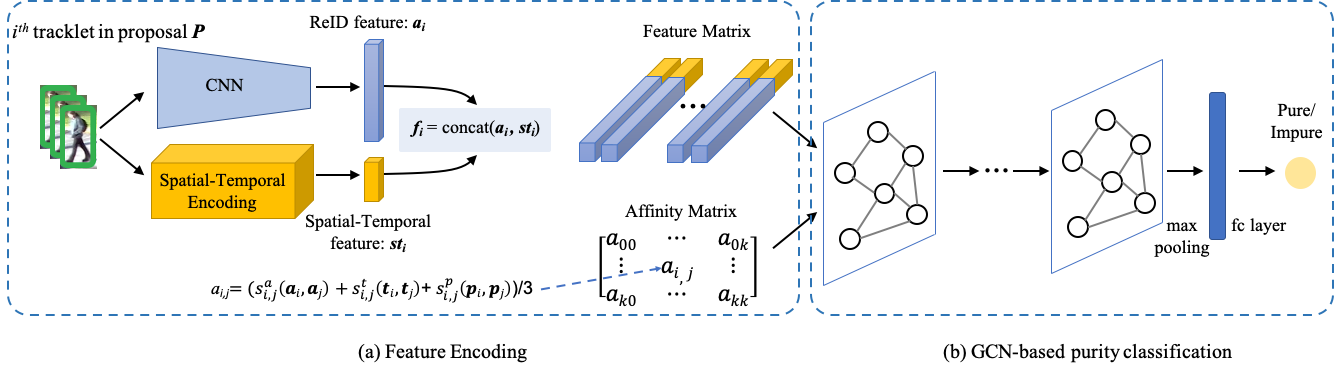}
\caption{Visualization of (a) feature encoding and (b) GCN-based purity classification netowrk.}
\label{fig:picture002}
\end{figure*}

We propose an iterative clustering strategy to grow the proposals gradually, as shown in Figure~\ref{fig:picture001}. It mainly consists of two modules.
%Algorithm~\ref{alg:IterProGen} shows the detailed procedure to produce proposals with $I$ iterations. 

\textbf{Affinity Graph Construction.}
At each iteration $i$, we build an affinity graph $\mathcal{G}$ to model the similarity between vertices $\mathcal{V} :=\{{v}_{1}, \cdots, {v}_{n} \}$.
Let vertex ${v}_{i}=(\mathbf{a}_{i}, \mathbf{t}_{i}, \mathbf{p}_{i})$, where $\mathbf{a}_{i}$ be the averaged ReID feature of a proposal, $\mathbf{t}_{i}$=$[t_{i}^{s}, \dots, t_{i}^{e}]$ be the sorted timestamps of detections in the proposal, $\mathbf{p}_{i}$=$[p_{i}^{s}, \dots, p_{i}^{e}]$ be the corresponding 2D image coordinates.
The affinity score of an edge (${v}_{i}$, ${v}_{j}$) is defined as the average score based on temporal, spatial and appearance similarities.
\begin{align}
  a_{ij}({v}_{i}, {v}_{j}) &= \frac{1}{3}(s_{ij}^{a}(\mathbf{a}_{i},\mathbf{a}_{j}) +  s_{ij}^{t}(\mathbf{t}_{i},\mathbf{t}_{j}) +  s_{ij}^{p}(\mathbf{p}_{i},\mathbf{p}_{j})) \label{equa:affinity} \\
  s_{ij}^{a}(\mathbf{a}_{i},\mathbf{a}_{j}) &= \frac{\mathbf{a}_{i} \cdot \mathbf{a}_{j}}{| \mathbf{a}_{i} | \cdot | \mathbf{a}_{j} |} \label{equa:appsimi} \\
  s_{ij}^{t}(\mathbf{t}_{i},\mathbf{t}_{j}) &=\begin{cases}
   exp(-\frac{\mathbf{g}(\mathbf{t}_{i}, \mathbf{t}_{j})}{\sigma_{t}}), &if \ \ \mathbf{g}(\mathbf{t}_{i}, \mathbf{t}_{j})> 0  \\
   -inf, &otherwise  \label{equa:apptt}
\end{cases} \\
  s_{ij}^{p}(\mathbf{p}_{i},\mathbf{p}_{j}) &= exp(-\frac{{f}(\mathbf{p}_{i}, \mathbf{p}_{j})}{\sigma_{p}}) \label{equa:apppos}
\end{align}
where $\mathbf{g}(\cdot)$ measures the minimum time gap between two vertices and  $\mathbf{g}(\mathbf{t}_{i}, \mathbf{t}_{j})$ = -1 if vertex ${v}_{i}$ has temporal overlapping with vertex ${v}_{j}$, ${f}(\cdot)$ measures the Euclidean distance between the predicted box~\footnote{We apply a global constant velocity model to predict the 2D image coordinates of the bounding box.} center of vertex ${v}_{i}$ and the starting box center of vertex ${v}_{j}$, $\sigma_{t}$ and $\sigma_{p}$ are controlling parameters.
%$\sigma_{t}$ and $\sigma_{p}$ are the parameters controlling the sensitivity of the temporal and spatial affinity score, respectively.
To reduce the complexity of the graph, a simple gating strategy is adopted (see Appendix A.1 for details) and the maximum number of edges linked to one vertex is set to be less than $K$.
%It means that only top-$K$ neighbors with highest affinity scores will be kept for each vertex among all valid neighbors defined by Eq.~\ref{equa:gatestrategy}.

\textbf{Cluster Proposals.}
%Algorithm 2 in Appendix A shows the detailed procedure to cluster nodes in the graph.
The basic idea of proposal generation is to use connected components to find clusters.
In order to keep the purity of the generated clusters high in the early iterations, we constrain the maximum size of each cluster to be below a threshold $s_{max}$.
In this phase, the vertices of a target object may be over-fragmented into several clusters.
%Note that we also adopt a function to keep all pairwise vertices within a cluster to be temporally compatible, i.e., no temporally overlapping vertices are allowed within same cluster.
The clusters generated in iteration $i$ are used as the input vertices of the next iteration.
And a new graph can be built on top of these clusters, thereby producing clusters of larger sizes.
%As the number of iterations increases, the parameters are gradually relaxed to grow proposals further, thereby increasing the recall rates.
%In the beginning phase, a set of very conservative parameters are used to keep the purity of the generated proposals. 
%In this phase, the vertices of a target object may be over-fragmented into several clusters. As the number of iterations increases, the parameters are gradually relaxed to grow proposals further, thereby increasing the recall rates. 
The final proposal set includes all the clusters in each iteration, thus providing an over-complete and diverse set of proposals $\mathcal{P} = \{ \mathcal{P}_{i} \}$.
%With this new graph, we can apply proposal generation again and obtain clusters of larger sizes.
%The union of grouped clusters in each iteration becomes the final proposal set $\mathcal{P} = \{ \mathcal{P}_{i} \}$.
The exact procedures are detailed in Algorithm 1 and 2 in Appendix A.2.
%With multiple iterations of Algorithm~\ref{alg:IterProGen}, we can obtain proposals at multiple scales $\mathcal{P} = \{ \mathcal{P}_{i} \}$.

% WONGUN reviewed up to this.... 2020/09/27

\subsection{Purity Classification Network}\label{subsectionPCN}

In this subsection, we devise the purity classification network to estimate the precision scores $\{ prec(\mathcal{P}_{i}) \}$ of the generated proposals $\mathcal{P}$.
%Assume that pure proposals usually exhibit certain structural patterns among vertices.
%To this end, a GCN is introduced to identify such patterns, i.e., to differentiate between pure and impure proposals.
Specifically, given a proposal $\mathcal{P}_{i}$ = $\{ {v}_{i} \}_{i=1}^{N_{i}}$ with $N_{i}$ vertices, the GCN takes the features associated with its vertices and sub-graph affinity matrix as input and predicts the probability of $\mathcal{P}_{i}$ being pure.
As shown in Figure~\ref{fig:picture002}, this module consists of the following two main parts. 
%one is feature encoding, the other is GCN-based classifier.

\textbf{Design of Feature Encoding.} 
Both the appearance and the spatial-temporal features are crucial cues for MOT.
%As shown in Figure~\ref{fig:picture002} (a), 
For appearance features, a CNN is applied to extract feature embeddings $a_{i}$ directly from RGB data of each detection $d_{i}$.
Then, we obtain ${v}_{i}$'s corresponding appearance features $\mathbf{a}_{i}$ by taking the average value of all detection appearance features.
%A vertex ${v}_{i}$ (i.e., a track-let) in each proposal is composed of a set of detections.
%We obtain ${v}_{i}$'s corresponding appearance features $\mathbf{a}_{i}$ by taking the average value of all detection appearance features.
For spatial-temporal features, we seek to obtain a representation that encodes, for each pair of temporal adjacent tracklets, their relative position, relative box size, as well as distance in time.
For proposal $\mathcal{P}_{i}$ = $\{ {v}_{i} \}_{i=1}^{N_{i}}$, its vertices are sorted first in ascending order according to the start timestamp of each vertex.
Then, for every pair of temporal adjacent tracklets ${v}_{i}$ and ${v}_{i+1}$, the ending timestamp of ${v}_{i}$ and the starting timestamp of ${v}_{i+1}$ is denoted as $t_{e_{i}}$ and $t_{s_{i+1}}$ respectively. And their bounding box coordinates in these timestamps are parameterized by top left corner image coordinates,  width and height, i.e., ($x_{i}$, $y_{i}$, $w_{i}$, $h_{i}$) and ($x_{i+1}$, $y_{i+1}$, $w_{i+1}$, $h_{i+1}$).
We compute the spatial-temporal feature $\mathbf{st}_{i}$ for vertex ${v}_{i}$ as:
\begin{equation}\label{equa:spatemfea}
    (\frac{2(x_{i+1} - x_{i})}{w_{i} + w_{i+1}}, \frac{2(y_{i+1} - y_{i})}{h_{i} + h_{i+1}}, log\frac{h_{i+1}}{h_{i}}, log\frac{w_{i+1}}{w_{i}}, t_{s_{i+1}} - t_{e_{i}} )
\end{equation}
if $i > 0$ else $\mathbf{st}_{i} = (1, 0, 0, 0, 0)$. With appearance feature $\mathbf{a}_{i}$ and spatial-temporal feature $\mathbf{st}_{i}$ at hand, we concatenate them to form the feature encoding $\mathbf{f}_{i}$ = $concat(\mathbf{a}_{i}, \mathbf{st}_{i})$ for each vertex ${v}_{i}$.

\textbf{Design of GCN.} 
As described above, we have obtained the features associated to vertices in $\mathcal{P}_{i}$ (denoted as $\mathbb{F}_{0}(\mathcal{P}_{i})$).
As for the affinity matrix for $\mathcal{P}_{i}$ (denoted as $\mathbb{A}(\mathcal{P}_{i})$), a fully-connected graph is adopted, in which we compute the affinity between each pair of vertices, as shown in Figure \ref{fig:picture002} (a).
The GCN network consists of $L$ layers and the computation of each layer can be formulated as:
\begin{equation}\label{equa:GCNlayer}
  \mathbb{F}_{l+1}(\mathcal{P}_{i}) = \sigma(\mathbb{D}(\mathcal{P}_{i})^{-1}\cdot (\mathbb{A}(\mathcal{P}_{i}) + \mathbb{I})\cdot \mathbb{F}_{l}(\mathcal{P}_{i})\cdot \mathbb{W}_{l})
\end{equation}
where $\mathbb{D}(\mathcal{P}_{i})$ = $\sum_{j}\mathbb{A}_{ij}(\mathcal{P}_{i})$ is the diagonal degree matrix. $\mathbb{F}_{l}(\mathcal{P}_{i})$ indicates the feature embeddings of the $l$-th layer, $\mathbb{W}_{l}$ represents the transform matrix, and $\sigma$ is a non-linear activation function ($ReLU$ in our implementation).
At the top-level feature embedding $\mathbb{F}_{L}(\mathcal{P}_{i})$, a max pooling is applied over all vertices in $\mathcal{P}_{i}$ to provide an overall summary.
Finally, a fully-connected layer is employed to classify $\mathcal{P}_{i}$ into a pure or impure proposal.
As shown in Equation~\ref{equa:GCNlayer}, for each GCN layer, it actually does three things: 
1) computes the weighted average of the features of each vertex and its neighbors;
2) transforms the features with $\mathbb{W}_{l}$;
3) feeds the transformed features to a nonlinear activation function.
Through this formulation, the purity network can learn the inner  consistency  of proposal $\mathcal{P}_{i}$.

\begin{comment}
\textbf{Training and Inference.} 
During training phase, we prepare a set of training samples via proposal generation in the training dataset.
Each sample contains a feature matrix, an affinity matrix, as well as a training label to indicate whether this sample is pure or not.
We train the purity classification network by using the binary-cross-entropy as the loss function.
During inference, the generated proposals are fed to the trained purity classification network, and output the predicted results.
It should be noticed that we use a threshold of 0.5 to binarize the inference results, i.e., if the inference result is larger than 0.5, we set it as 1, otherwise 0.
\end{comment}

\subsection{Trajectory Inference}\label{subsectionDO}
With the purity inference results, we can obtain the quality scores of all proposals with Equation~\ref{equa:quality_score}.
A simple de-overlapping algorithm is adopted to guarantee that each tracklet is assigned one unique track ID.
%In order to guarantee that each track-let is assigned one unique track ID, a trajectory inference step is needed.
%A simple de-overlapping algorithm is adopted in this paper.
First, we rank the proposals in descending order of the quality scores.
Then, we sequentially assign track ID to vertices in the proposals from the ranked list, and modify each proposal by removing the vertices seen in preceding ones. The detailed algorithm is described in Algorithm 3 in Appendix A.2.

\section{Experiments} \label{section:experiment}
In this section, we first present an ablation study to better understand the behavior of each module in our pipeline. Then, we compare our methods to published methods on the MOTChallenge benchmarks.
\subsection{Experimental Setup}

\subsubsection{Datasets and metrics}
All experiments are done on the multiple object tracking benchmark MOTChallenge, which consists of several challenging pedestrian tracking sequences with frequent occlusions and crowded scenes.
We choose two separate tracking benchmarks, namely MOT17~\cite{milan2016mot16} and MOT20~\cite{dendorfer2020mot20}.
These two benchmarks consist of challenging video sequences with varying viewing angle, size, number of objects, camera motion, illumination and frame rate in unconstrained environments.
To ensure a fair comparison with other methods, we use the public detections provided by MOTChallenge, and preprocess them by first running~\cite{bergmann2019tracking}. 
This strategy is widely used in published methods~\cite{braso2020learning, inproceedingsLiu}.

For the performance evaluation, we use the widely accepted MOT metrics~\cite{bernardin2008evaluating, wu2006tracking, ristani2016performance}, including Multiple Object Tracking Accuracy (MOTA), ID F1 score (IDF1), Mostly Track targets (MT), Mostly Lost targets (ML), False Positives (FP), False Negatives (FN), ID switches (IDs), etc.
Among these metrics, MOTA and IDF1 are the most important ones, as they quantify two of the main aspects of MOT, namely, object coverage and identity preservation.

\subsubsection{Implementation details}
\textbf{ReID Model.} For the CNN network used to extract ReID features, we employ a variant of ResNet50, named ResNet50-IBN~\cite{luo2019strong}, which replaces batch norm layer with instance-batch-norm (IBN) layer. 
After global average pooling layer, a batch norm layer and a classifier layer is added.
We use triplet loss and ID loss to optimize the model weights.
For the ablation study, we use the ResNet50-IBN model trained on two publicly available datasets: ImageNet~\cite{deng2009imagenet} and Market1501~\cite{zheng2015scalable}.
While for the final benchmark evaluation, we add the training sequences in MOT17~\cite{milan2016mot16} and MOT20~\cite{dendorfer2020mot20} to finetune the ResNet50-IBN model.
Note that using training sequences in the benchmark to finetune ReID model for the test sequences is a common practice among MOT methods~\cite{hornakova2020lifted, kim2018multi, tang2017multiple}.

\textbf{Parameter Setting.} 
In affinity graph construction, the parameter $\sigma_{t}$ and $\sigma_{p}$ is empirically set to 40 and 100, respectively.
%All gating parameters (i.e., $\tau_{t}$, $\tau_{p}$, $\tau_{a}$) are automatically determined by computing the calibration information in the training set.
In proposal generation, the maximum iteration number is set to $I$=10, the maximum neighbors for each node is set to $K$=3, the maximum cluster size is set to $s_{max}$=2, and the cluster threshold step is set to $\Delta$=0.05.
In trajectory inference, the weighting parameter $w$ is set to 1 and $C$=200.

\textbf{GCN Training.} 
We use a GCN with $L$=4 hidden layers in our experiments.
The GCN model is trained end-to-end with Adam optimizer, where weight decay term is set to $10^{-4}$, $\beta_{1}$ and $\beta_{2}$ is set to 0.9 and 0.999, respectively.
The batch size is set to 2048.
We train for 100 iterations in total with a learning rate $10^{-3}$.
For data augmentation, we randomly remove detections to simulate missed detections.
For the ablation study, the leave-one-out cross-validation strategy is adopted to evaluate the GCN model.

\textbf{Post Processing.} 
We perform simple bilinear interpolation along missing frames to fill gaps in our trajectories.

\subsection{Ablation Study}
In this subsection, we aim to evaluate the performance of each module in our framework. We conduct all of our experiments with the training sequences of the MOT17 datasets.

\subsubsection{Proposal Generation} 
To evaluate the performance of proposal generation, we choose the oracle purity network for proposal purity classification, i.e., determine whether the proposal $\mathcal{P}_{i}$ is pure or not by comparing it with the ground-truth data. 
%And a simple de-overlapping strategy shown in Algorithm~\ref{alg:deoverlapping} is adopted for inference.
For baseline, we adopt the MHT algorithm \cite{kim2015multiple} by removing the $N$-scan prunning step.
To reduce the search space, a simple gating strategy is adopted which limits the maximum number of linkage for each vertex to be less than 20.
The comparison results are summarized in Table \ref{tab:pro_eval}.
As expected, the time cost of our iterative proposal generation method is far less than that of the MHT-based method.
Meanwhile, our method can achieve comparable MOTA and IDF1 scores.
This demonstrates its ability to reduce the computational cost while guarantee the quality of the generated proposals.

\begin{table}[tbp]
\footnotesize
\centering
 \setlength{\tabcolsep}{0.9mm}{
 \begin{tabular}{lccccccccc} 
  \toprule 
  Alg.  & MOTA$\uparrow$ & IDF1$\uparrow$ & MT$\uparrow$ & ML$\downarrow$ & FP$\downarrow$ & FN$\downarrow$ & IDs$\downarrow$ & Hz$\uparrow$\\ 
  \midrule 
   Ours & \textbf{64.8} & 73.3 & \textbf{631} & \textbf{384} & 4006 & \textbf{113769} & 749 & \textbf{21.6} \\ 
   MHT & 64.7 & \textbf{73.6} & 632 & 389 & \textbf{3767} & 114495 & \textbf{608} & 2.4 \\ 
  \bottomrule 
 \end{tabular}}
 \\[10pt]
  \caption{\label{tab:pro_eval}Performance comparison with different proposal generation algorithms.}
\end{table}

\begin{figure}
\centering
\includegraphics[width=0.40\textwidth]{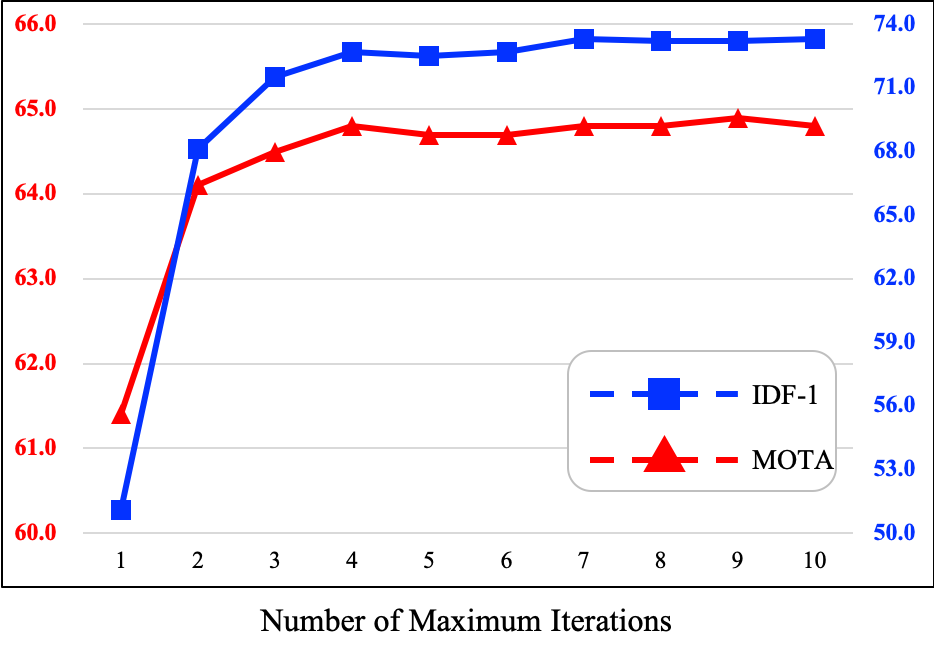}
\caption{Influence of the iteration number $I$ on proposal generation performance.}
\label{fig:picture003}
\end{figure}

\textbf{Effect of Maximum Iteration Number.} 
There are four parameters in proposal generation, namely $I$, $K$, $s_{max}$ and $\Delta$. Experimental results show that the tracking performance is insensitive to $K$, $s_{max}$ and $\Delta$. The detailed results are shown in Appendix B.
Intuitively, increasing the maximum iteration number $I$ allows to generate a larger number of proposals, and improves the possibility of the generated proposals to contain good tracklets under long-term occlusions.
Hence, one would expect higher $I$ values to yield better performance.
We test this hypothesis in Figure~\ref{fig:picture003} by doing proposal generation with increasing number of $I$, from 1 to 10.
As expected, we see a clear upward tendency for both MOTA and IDF1 metrics.
Moreover, it can be observed that the performance boost in both metrics mainly occurs when increasing $I$ from 1 to 2, which demonstrates that most of the occlusions are short-term.
We also observe that the upwards tendency for both MOTA and IDF1 metrics stagnates around seven iterations. 
There is a trade-off between performance and computational cost in choosing the proper number of iterations.
Hence, we use $I=10$ in our final configuration.

\subsubsection{Purity Classification Network}
\textbf{Effects of the features.}
Our GCN-based purity classification network receives two main streams of features for each vertex: (i) appearance features from ReID model, and (ii) spatial-temporal features from Equation~\ref{equa:spatemfea}.
We test their effectiveness by experimenting with combinations of the above two groups of features.
Results are summarized in Table~\ref{tab:feature_eval}.
It can be concluded that: (i) the appearance features seems to play a more important role in identity preservation, hence having higher IDF1 and MT measures, (ii) the spatial-temporal features can reduce the the number of FP and IDs, and (iii) combination of these two streams of features can improve the overall performance.

\textbf{Effects of different loss functions.}
We perform an experiment to study the impact of different loss functions in model training.
Table~\ref{tab:lossfunc_eval} lists the detailed quantitative comparison results by using binary-cross-entropy loss (BCELoss) and mean-squared-error loss (MSELoss), respectively.
Using BCELoss shows a gain of 0.6 IDF1 measure and a small amount of decrease of IDs.
Hence, we use BCELoss in our final configuration.

\begin{table}[tbp]
\footnotesize
\centering
 \setlength{\tabcolsep}{0.9mm}{
 \begin{tabular}{lcccccccc} 
  \toprule 
  Feats.  & MOTA$\uparrow$ & IDF1$\uparrow$ & MT$\uparrow$ & ML$\downarrow$ & FP$\downarrow$ & FN$\downarrow$ & IDs$\downarrow$ \\ 
  \midrule
   Spat+Temp & 63.6 & 69.4 & 622 & 381 & \textbf{5152} & 116653 & 819  \\
   App & \textbf{64.0} & 70.3 & 634 & \textbf{373} & 6297 & 113865 & 1076  \\
   Spat+Temp+App & 63.9 & \textbf{71.8} & \textbf{647} & 377 & 7176 & \textbf{113700} & \textbf{728}  \\
  \bottomrule 
 \end{tabular}}
 \\[10pt]
  \caption{\label{tab:feature_eval}Performance comparison for GCN-based purity classification network with different features.} 
\end{table}

\begin{table}[tbp]
\footnotesize
\centering
 \setlength{\tabcolsep}{0.9mm}{
 \begin{tabular}{lcccccccc} 
  \toprule 
  Training Loss  & MOTA$\uparrow$ & IDF1$\uparrow$ & MT$\uparrow$ & ML$\downarrow$ & FP$\downarrow$ & FN$\downarrow$ & IDs$\downarrow$ \\ 
  \midrule 
   BCELoss & \textbf{63.9} & \textbf{71.8} & \textbf{647} & \textbf{377} & \textbf{7176} & \textbf{113700} & \textbf{728}  \\
   MSELoss & 63.8 & 71.2 & 646 & 378 & 7422 & 113878 & 765  \\
  \bottomrule 
 \end{tabular}}
 \\[10pt]
  \caption{\label{tab:lossfunc_eval}Performance comparison for GCN-based purity classification network with different loss functions.} 
\end{table}

\begin{table}[tbp]
\footnotesize
\centering
 \setlength{\tabcolsep}{1.0mm}{
 \begin{tabular}{lcccccccc} 
  \toprule 
  Alg.  & MOTA$\uparrow$ & IDF1$\uparrow$ & MT$\uparrow$ & ML$\downarrow$ & FP$\downarrow$ & FN$\downarrow$ & IDs$\downarrow$ \\ 
  \midrule 
   Oracle & \textbf{64.8} & \textbf{73.3} & 631 & 384 & \textbf{4006} & 113769 & 749  \\ 
   GCN & 63.9 & 71.8 & \textbf{647} & 377 & 7176 & \textbf{113700} & \textbf{728}  \\
   TCN & 63.8 & 70.6 & 628 & 379 & 6510 & 114666 & 901  \\
   ALSTM & 63.5 & 69.5 & 634 & 380 & 6131 & 115756 & 1045  \\
   ALSTM-FCN  & 63.7 & 69.4 & 621 & \textbf{373} & 4897 & 116354 & 1087  \\
  \bottomrule 
 \end{tabular}}
 \\[10pt]
  \caption{\label{tab:network_eval}Performance comparison with different purity classification networks.} 
\end{table}

\textbf{Effects of different networks.}
There are numerous previous works that use deep neural networks, such as Temporal Convolutional Network (TCN~\cite{bai2018empirical}), Attention Long-Short Term Memory (ALSTM~\cite{karim2019multivariate}), ALSTM Fully Convolutional Network (ALSTM-FCN~\cite{karim2019multivariate}) to conduct temporal reasoning on the sequence of observations.
Table~\ref{tab:network_eval} presents the results by using these neural networks.
It should be noticed that the oracle performance in Table~\ref{tab:network_eval} is obtained by using ground-truth data for purity classification.
By comparing GCN with Oracle, we can see that GCN obtains better MT and ML measures, but worse MOTA and IDF1 measures than Oracle.
The reason might be due to the false positives in GCN-based proposal purity classification, which would generate a few impure trajectories and hence reduce IDF1 measure.
Moreover, the impure trajectories would cause quite a few FPs in the post processing (as shown in Table~\ref{tab:network_eval}), hence reducing the MOTA measure.
By comparing GCN with other neural networks, it is clear that GCN achieves better performance on most metrics, improving especially the IDF1 measure by 1.2 percentage.
The performance gain is attributed to its capability of learning higher-order information in a message-passing way to measure the purity of each proposal.
It verifies that GCN is more suitable for solving the proposal classification problem.

\subsubsection{Trajectory Inference}
The iterative greedy strategy is a widely used technique in MOT, which can be an alternative choice of inference.
Specifically, it iteratively performs the following steps: 
first, estimate the quality scores of all existing proposals; 
second, collect the proposal with highest quality score and assign unique track ID to the vertices within this proposal; 
third, modify the remaining proposals by removing the vertices seen in preceding ones.
Hence, the computational complexity of the iterative greedy strategy is $O(N^{2})$.
Compared with the iterative greedy strategy, the simple de-overlapping algorithm only estimates the quality scores once.
Therefore, it can reduce the computational complexity to $O(N)$.
The comparison results are summarized in Table~\ref{tab:deoverlapping_eval}.
It can be observed that the simple de-overlapping algorithm achieves slightly better performance in both MOTA and IDF1 metrics than the iterative greedy strategy.
The reason might be due to that as the number of iteration increases, the number of nodes in each proposal decreases.
Hence, the classification accuracy of the purity network might decrease.

\begin{table}[tbp]
\footnotesize
\centering
 \setlength{\tabcolsep}{0.9mm}{
 \begin{tabular}{lcccccccc} 
  \toprule 
  De-overlapping  & MOTA$\uparrow$ & IDF1$\uparrow$ & MT$\uparrow$ & ML$\downarrow$ & FP$\downarrow$ & FN$\downarrow$ & IDs$\downarrow$ \\ 
  \midrule
   Simple & \textbf{63.9} & \textbf{71.8} & \textbf{647} & \textbf{377} & \textbf{7176} & 113700 & 728  \\
   Iterative Greedy & 63.6 & 71.7 & \textbf{647} & \textbf{377} & 8628 & \textbf{113449} & \textbf{719}  \\
  \bottomrule 
 \end{tabular}}
 \\[10pt]
  \caption{\label{tab:deoverlapping_eval}Performance comparison with different de-overlapping strategies.} 
\end{table}

\subsection{Benchmark Evaluation}
We report the quantitative results obtained by our method on MOT17 and MOT20 in Table~\ref{tab:mot17_eval} and Table~\ref{tab:mot20_eval} respectively, and compare it to methods that are officially published on the MOTChallenge benchmark.
As shown in Table~\ref{tab:mot17_eval} and Table~\ref{tab:mot20_eval}, our method obtains state-of-the-art results, improving especially the IDF1 measure by 1.2 percentage points on MOT17 and 3.4 percentage points on MOT20.
It demonstrates that our method can achieve strong performance in identity preservation.
We attribute this performance increase to our proposal-based learnable framework.
First, our proposal generation module generates an over-complete set of proposals, which improves its anti-interference ability in challenging scenarios such as occlusions.
Second, our GCN-based purity network directly optimizes the whole proposal score rather than the pairwise matching cost, which takes higher-order information into consideration to make globally informed predictions.
We also provide more comparison results with other methods on MOT16~\cite{milan2016mot16} benchmark in Appendix C.

Our method outperforms MPNTrack~\cite{braso2020learning} only by a small margin in terms of the MOTA score.
It should be noticed that MOTA measures the object coverage and overemphasizes detection over association~\cite{luiten2020hota}.
%is mainly determined by the performance of the detector.
We use the same set of detections and post-processing strategy (simple bilinear interpolation) as MPNTrack~\cite{braso2020learning}.
Then, achieving similar MOTA results is in line with expectations.
IDF1 is preferred over MOTA for evaluation due to its focus on measuring association accuracy over detection accuracy.
We also provide more qualitative results in Appendix D.

\begin{table}[tbp]
\footnotesize
\centering
\begin{threeparttable}
 \setlength{\tabcolsep}{0.8mm}{
 \begin{tabular}{lccccccccc} 
  \toprule 
  Method  & MOTA$\uparrow$ & IDF1$\uparrow$ & MT$\uparrow$ & ML$\downarrow$ & FP$\downarrow$ & FN$\downarrow$ & IDs$\downarrow$  & Hz$\uparrow$ \\ 
  \midrule
   Ours & 59.0 & \textbf{66.8} & \textbf{29.9} & 33.9 & 23102 & 206948 & \textbf{1122} & 4.8   \\
   Lif\_T\cite{hornakova2020lifted} & \textbf{60.5} & 65.6 & 27.0 & 33.6 & 14966 & \textbf{206619} & 1189 & 0.5   \\
   MPNTrack\cite{braso2020learning} & 58.8 & 61.7 & 28.8 & 33.5 & 17413 & 213594 & 1185 & \textbf{6.5}  \\
   JBNOT\cite{henschel2019multiple} & 52.6 & 50.8 & 19.7 & 35.8 & 31572 & 232659 & 3050 & 5.4  \\
   eHAF\cite{sheng2018heterogeneous} & 51.8 & 54.7 & 23.4 & 37.9 & 33212 & 236772 & 1834 & 0.7  \\
   NOTA\cite{chen2019aggregate} & 51.3 & 54.7 & 17.1 & 35.4 & 20148 & 252531 & 2285 & -  \\
   FWT\cite{henschel2017improvements} & 51.3 & 47.6 & 21.4 & 35.2 & 24101 & 247921 & 2648 & 0.2  \\
   jCC\cite{keuper2018motion} & 51.2 & 54.5 & 20.9 & 37.0 & 25937 & 247822 & 1802 & 1.8 \\
   \hline
   GNNMatch\cite{papakis2020gcnnmatch} & 57.3 & 56.3 & 24.2 & \textbf{33.4} & 14100 & 225042 & 1911 & 1.3 \\
   Tracktor\cite{bergmann2019tracking} & 56.3 & 55.1 & 21.1 & 35.3 & \textbf{8866} & 235449 & 1987 & 1.8  \\
   FAMNet\cite{chu2019famnet} & 52.0 & 48.7 & 19.1 & \textbf{33.4} & 14138 & 253616 & 3072 & -  \\
  \bottomrule 
 \end{tabular}}
 \vspace{10pt}
  \caption{\label{tab:mot17_eval}Performance comparison with start-of-the art on MOT17  (top: offline methods; bottom: online methods).} 
  \end{threeparttable}
\end{table}

\begin{table}[tbp]
\footnotesize
\centering
\begin{threeparttable}
 \setlength{\tabcolsep}{0.65mm}{
 \begin{tabular}{lccccccccc} 
  \toprule 
  Method  & MOTA$\uparrow$ & IDF1$\uparrow$ & MT$\uparrow$ & ML$\downarrow$ & FP$\downarrow$ & FN$\downarrow$ & IDs$\downarrow$  & Hz$\uparrow$ \\ 
  \midrule
   Ours & 56.3 & \textbf{62.5} & 34.1 & 25.2 & 11726 & 213056 & 1562 & 0.7   \\
   MPNTrack\cite{braso2020learning} & \textbf{57.6} & 59.1 & \textbf{38.2} & \textbf{22.5} & 16953 & \textbf{201384} & \textbf{1210} & 6.5  \\
   \hline
   GNNMatch\cite{papakis2020gcnnmatch} & 54.5 & 49.0 & 32.8 & 25.5 & 9522 & 223611 & 2038 & 0.1 \\
   UnsupTrack~\cite{karthik2020simple} & 53.6 & 50.6 & 30.3 & 25.0 & \textbf{6439} & 231298 & 2178 & 1.3 \\
   SORT20~\cite{bewley2016simple} & 42.7 & 45.1 & 16.7 & 26.2 & 27521 & 264694 & 4470 & \textbf{57.3} \\
  \bottomrule 
 \end{tabular}}
 \vspace{10pt}
  \caption{\label{tab:mot20_eval}Performance comparison with start-of-the art on MOT20.} 
  \end{threeparttable}
\end{table}

\section{Conclusion} \label{section:conclusion}
In this paper, we propose a novel proposal-based MOT learnable framework.
%which formulates MTT as a proposal generation, proposal scoring and trajectory inference paradigm on an affinity graph.
For proposal generation, we propose an iterative graph clustering strategy which strikes a good trade-off between proposal quality and computational cost.
For proposal scoring, a GCN-based purity network is deployed to capture  higher-order information within each proposal, hence improving anti-interference ability in challenge scenarios such as occlusions.
We experimentally demonstrate that our method achieves a clear performance improvement with respect to previous state-of-the-art.
For future works, we plan to make our framework be trainable end-to-end especially for the task of proposal generation.

%\section*{Acknowledgements}
\textbf{Acknowledgements.} This research is funded by the National Key Research and Development Program of China (No. 2018AAA0100701)

{\small
\bibliographystyle{ieee_fullname}
\bibliography{egbib}
}

\newpage
\appendix
\appendixpage
\addappheadtotoc
\section{Detailed Algorithm}\label{section:supplementary_material_A}
In this section, we first detail the gating strategy in affinity graph construction, and then provide the pseudocode of the algorithms presented in the main paper.

\subsection{Gating Strategy}
To reduce the complexity of the graph, we adopt a simple gating strategy to remove the edges exceeding the thresholds.
Specifically, let $\mathcal{O}_{i}$ represent the valid neighbors of vertex $\textbf{v}_{i}$, and $\mathcal{O}_{i}$ is obtained by:
\begin{equation}\label{equa:gatestrategy}
\mathcal{O}_{i} = \{ \forall v_j;\   
\mathcal{I}^t(\mathbf{t}_{i}, \mathbf{t}_{j}, \tau_{t}) \& 
\mathcal{I}^p(\mathbf{p}_{i}, \mathbf{p}_{j}, \tau_{p}) \& \mathcal{I}^a(\mathbf{a}_{i}, \mathbf{a}_{j}, \tau_{a}) \}
\end{equation}
where $\mathcal{I}^t$ is an indicator function to check if the minimum time gap between vertex $\textbf{v}_{i}$ and $\textbf{v}_{j}$ is less than $\tau_{t}$, $\mathcal{I}^p$ is also an indicator function to check if the location distance is less than $\tau_{p}$ when having the minimum time gap, and $\mathcal{I}^a$ checks if the appearance distance is less than $\tau_{a}$.
The thresholds $\tau_{t}$, $\tau_{p}$ and $\tau_{a}$ determine the radius of the gate.

\subsection{Proposal Generation and Deoverlapping}
Algorithm~\ref{alg:IterProGen} and Algorithm~\ref{alg:ClusterNodes} show the detailed procedures to generate proposals. 
In these algorithms, $s_{max}$ (maximum cluster size) and $\Delta$ (cluster threshold step) are utilized to improve the purity of the generated clusters in the early iterations.
It should be noted that we adopt a compatible function to keep all pairwise vertices within a cluster to be temporally compatible, i.e., no temporally overlapping vertices are allowed within the same cluster.

Algorithm~\ref{alg:deoverlapping} provides a summary of the de-overlapping procedures to generate the final tracking output. 

\begin{algorithm}[htb]
\caption{Iterative Proposal Generation}\label{alg:IterProGen}
\KwIn{ Node set $\mathcal{V}$, iterative number $I$, maximum cluster size $s_{max}$, cluster threshold step $\Delta$.}
\KwOut{ Proposal set $\mathcal{P}$}
%\begin{algorithmic}
\textbf{initialization:} $\mathcal{P}=\varnothing$, $i$ = 0, $\mathcal{V}'=\mathcal{V}$\\
 \While{$i<I$}{
  $\mathcal{G}$ = $BuildAffinityGraph$($\mathcal{V}'$) \;
  $\mathcal{C}$ = $ClusterNodes$($\mathcal{G}$, $s_{max}$, $\Delta$) \;
  $\mathcal{P} = \mathcal{P} \cup \mathcal{C}$\;
  $\mathcal{V}'$ = $UpdateNodes$($\mathcal{C}$) \;
  $i$ = $i$ + 1 \;
 }
 \textbf{Return} $\mathcal{P}$
%\end{algorithmic}
\end{algorithm}

\begin{algorithm}
\caption{Cluster Nodes}\label{alg:ClusterNodes}
\KwIn{Symmetric affinity matrix $\mathcal{G}$, maximum cluster size $s_{max}$, cluster threshold step $\Delta$.}
\KwOut{ Clusters $\mathcal{C}$}
\SetKwFunction{FMain}{$main$}
\SetKwProg{Fn}{function}{:}{}
  \Fn{\FMain}{
    $\mathcal{C}=\varnothing$, $\mathcal{R}=\varnothing$, $\tau = min(\mathcal{G})$ \;
    $\mathcal{C}', \mathcal{R} = FindClucters(\mathcal{G}, \tau, s_{max})$ \;
    $\mathcal{C} = \mathcal{C} \cup \mathcal{C}'$ \;
    \While{$\mathcal{R} \neq \varnothing$}{
        $\tau =  \tau + \Delta $\;
        $\mathcal{C}', \mathcal{R} = FindClucters(\mathcal{G}_{\mathcal{R}}, \tau, s_{max})$ \;
        $\mathcal{C} = \mathcal{C} \cup \mathcal{C}'$ \;
    }
        \KwRet $\mathcal{C}$\;
  }
\SetKwFunction{FSub}{$FindClucters$}
\SetKwProg{Fn}{function}{:}{}
\Fn{\FSub{$\mathcal{G}, \tau, s_{max}$}}{
        $\mathcal{G}' = PruneEdge(\mathcal{G},  \tau)$ \;
        $\mathcal{S} = FindConnectedComponents(\mathcal{G}')$ \;
        $\mathcal{C}' = \{ c \left| c \in \mathcal{S}, \right. \left| c \right| < s_{max}$ and $Compatible(c) \}$ \;
        $\mathcal{R} = \mathcal{S} \backslash \mathcal{C}'$ \;
        \KwRet $\mathcal{C}', \mathcal{R}$\;
  }
\SetKwFunction{FSub}{$Compatible$}
\SetKwProg{Fn}{function}{:}{}
  \Fn{\FSub{$c$}}{
        \eIf{${\mathbf{d}}(\mathbf{t}_{i}, \mathbf{t}_{j}) > 0, \forall i, j \in c, i \neq j$}{
        \KwRet True \;}{
        \KwRet False \;} 
  }
\end{algorithm}

\begin{algorithm}[tb]
\caption{De-overlapping}\label{alg:deoverlapping}
\KwIn{Ranked Proposals $\{ \hat{\mathcal{P}}_{1},  \hat{\mathcal{P}}_{2}, \cdots , \hat{\mathcal{P}}_{N_{p}} \}$}
\KwOut{Tracking Results $\mathbb{T}$}
Dictionary $\mathbb{T}=\{ \}$, Occupied Set $\mathbf{I}=\varnothing, i=1$ \;
    \While{$i<=N_{p}$}{
        $\mathcal{C}_{i} = \hat{\mathcal{P}}_{i} \backslash \mathbf{I} $ \;
        \For{\texttt{ ${v}_i$ $in$ $\mathcal{C}_{i}$}} {
        \texttt{$\mathbb{T}[v_i]=i$} \;
        }
        $\mathbf{I} = \mathbf{I} \cup \mathcal{C}_{i}$ \;
        $i = i + 1$ \;
    }
Return $\mathbb{T}$ ;
\end{algorithm}

\section{Parameter Sensitivity Analysis} \label{section:supplementary_material_B}
\begin{figure}[tp]
\centering
\includegraphics[width=0.40\textwidth]{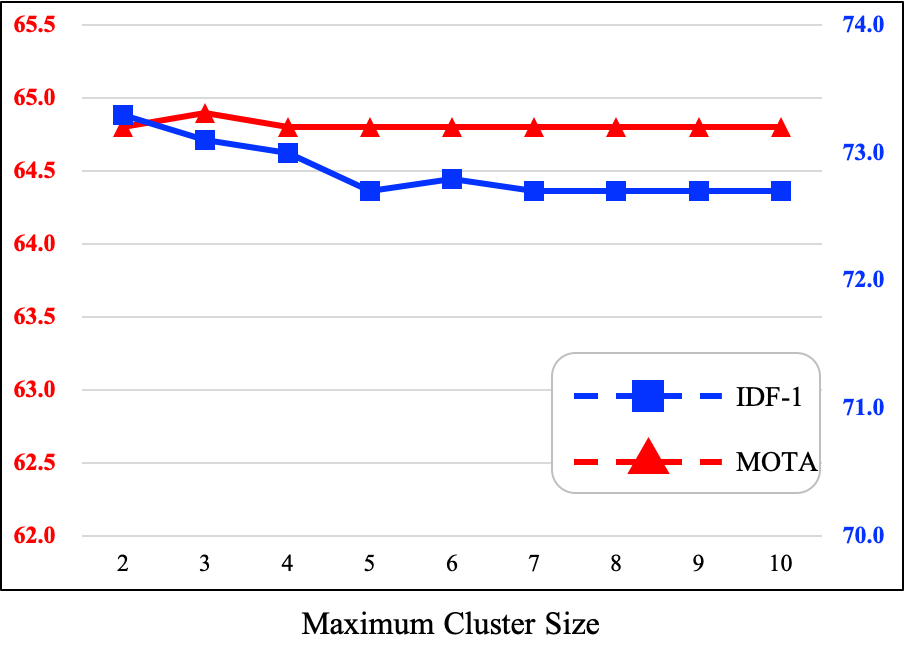}
\caption{Influence of the maximum cluster size $s_{max}$ on proposal generation performance.}
\label{fig:picture004}
\end{figure}

\begin{figure}[tp]
\centering
\includegraphics[width=0.40\textwidth]{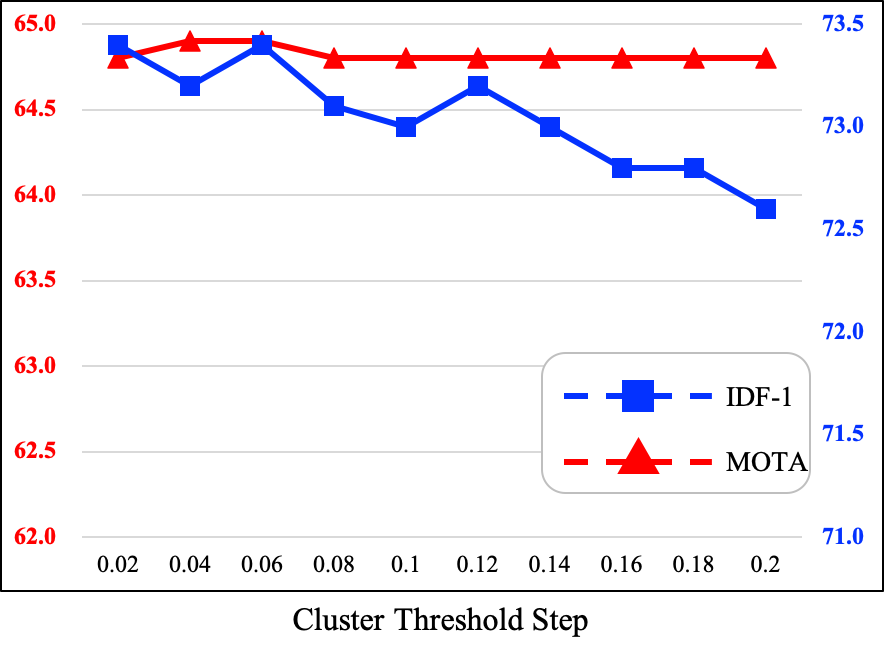}
\caption{Influence of the cluster threshold step $\Delta$ on proposal generation performance.}
\label{fig:picture005}
\end{figure}

\begin{figure}[tp]
\centering
\includegraphics[width=0.40\textwidth]{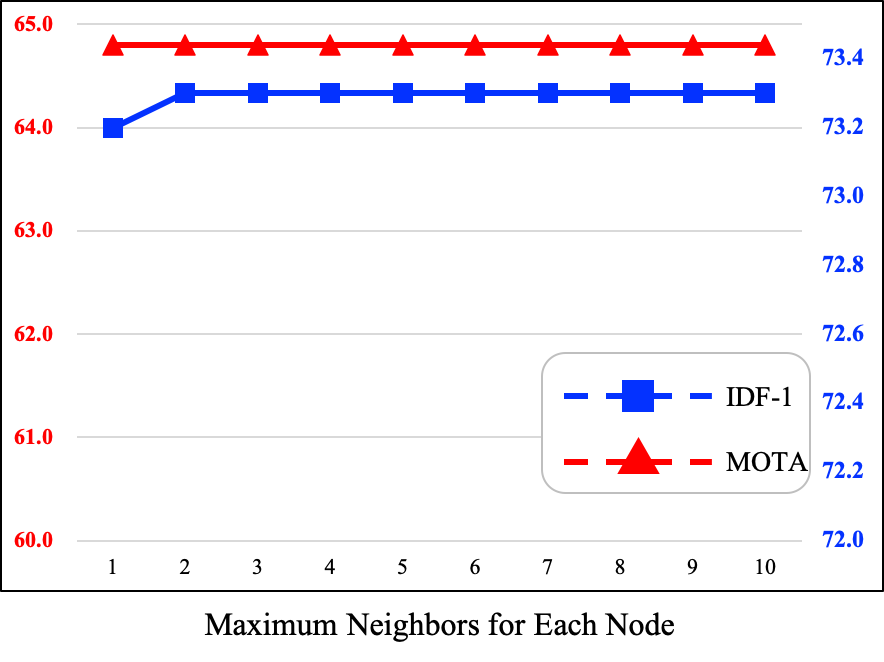}
\caption{Influence of the maximum neighbors for each node $K$ on proposal generation performance.}
\label{fig:picture006}
\end{figure}

Here, we investigate the effects of different settings on parameter $s_{max}$, $\Delta$ and $K$ (the maximum number of edges linked to one vertex) to the tracking performance.
The parameter $s_{max}$ and $\Delta$ are used to control the growth speed of the proposals. 
The results in Figure~\ref{fig:picture004} and Figure~\ref{fig:picture005} show that we can choose $s_{max} \in [2, 4]$, $\Delta \in [0.02, 0.06]$ to achieve the satisfactory and stable performance.
With the the increasing $s_{max}$ or $\Delta$, more noises will be introduced to the proposals in early iterations, hence reducing the performance.
The parameter $K$ controls the number of edges in the graph construction.
The results in Figure~\ref{fig:picture006} show that a satisfactory and stable performance can be achieved when $K>1$.

\section{Evaluation Results on MOT16}~\label{section:supplementary_material_C}
We also report the quantitative results obtained by our method on MOT16 in Table~\ref{tab:mot16_eval} and compare it to methods that are officially published on the MOTChallenge benchmark.
Our method can also obtain state-of-the-art IDF1 score on MOT16.

\begin{table}[tbp]
\footnotesize
\centering
\begin{threeparttable}
 \setlength{\tabcolsep}{0.8mm}{
 \begin{tabular}{lccccccccc} 
  \toprule 
  Method  & MOTA$\uparrow$ & IDF1$\uparrow$ & MT$\uparrow$ & ML$\downarrow$ & FP$\downarrow$ & FN$\downarrow$ & IDs$\downarrow$  & Hz$\uparrow$ \\ 
  \midrule
   Ours & 58.8 & \textbf{67.6} & \textbf{27.3} & 35.0 & 6167 & 68432 & 435 & 4.3   \\
   Lif\_T\cite{hornakova2020lifted} & 61.3 & 64.7 & 27.0 & 34.0 & 4844 & 65401 & 389 & 0.5   \\
   MPNTrack\cite{braso2020learning} & 58.6 & 61.7 & \textbf{27.3} & 34.0 & 4949 & 70252 & \textbf{354} & 6.5  \\
   HDTR\cite{babaee2018multiple} & 53.6 & 46.6 & 21.2 & 37.0 & 4714 & 79353 & 618 & 3.6   \\
   TPM\cite{peng2020tpm} & 51.3 & 47.9 & 18.7 & 40.8 & 2701 & 85504 & 569 & 0.8   \\
   CRF\_TRACK\cite{xiang2020end} & 50.3 & 54.4 & 18.3 & 35.7 & 7148 & 82746 & 702 & 1.5   \\
   NOTA\cite{chen2019aggregate} & 49.8 & 55.3 & 17.9 & 37.7 & 7248 & 83614 & 614 & \textbf{19.2}  \\
   \hline
   UnsupTrack\cite{karthik2020simple} & \textbf{62.4} & 58.5 & 27.0 & \textbf{31.9} & 5909 & \textbf{61981} & 588 & 1.9  \\
   GNNMatch\cite{papakis2020gcnnmatch} & 57.2 & 55.0 & 22.9 & 34.0 & 3905 & 73493 & 559 & 0.3 \\
   Tracktor\cite{bergmann2019tracking} & 56.2 & 54.9 & 20.7 & 35.8 & \textbf{2394} & 76844 & 617 & 1.6 \\
   TrctrD16\cite{xu2020train} & 54.8 & 53.4 & 19.1 & 37.0 & 2955 & 78765 & 645 & 1.6   \\
   PV\cite{li2019multi} & 50.4 & 50.8 & 14.9 & 38.9 & 2600 & 86780 & 1061 & 7.3   \\
  \bottomrule 
 \end{tabular}}
 \vspace{10pt}
  \caption{\label{tab:mot16_eval}Performance comparison with start-of-the art on MOT16  (top: offline methods; bottom: online methods).} 
  \end{threeparttable}
\end{table}

\section{Qualitative Analysis}~\label{section:supplementary_material_D}
\begin{figure}[tp]
\centering
\includegraphics[width=0.45\textwidth]{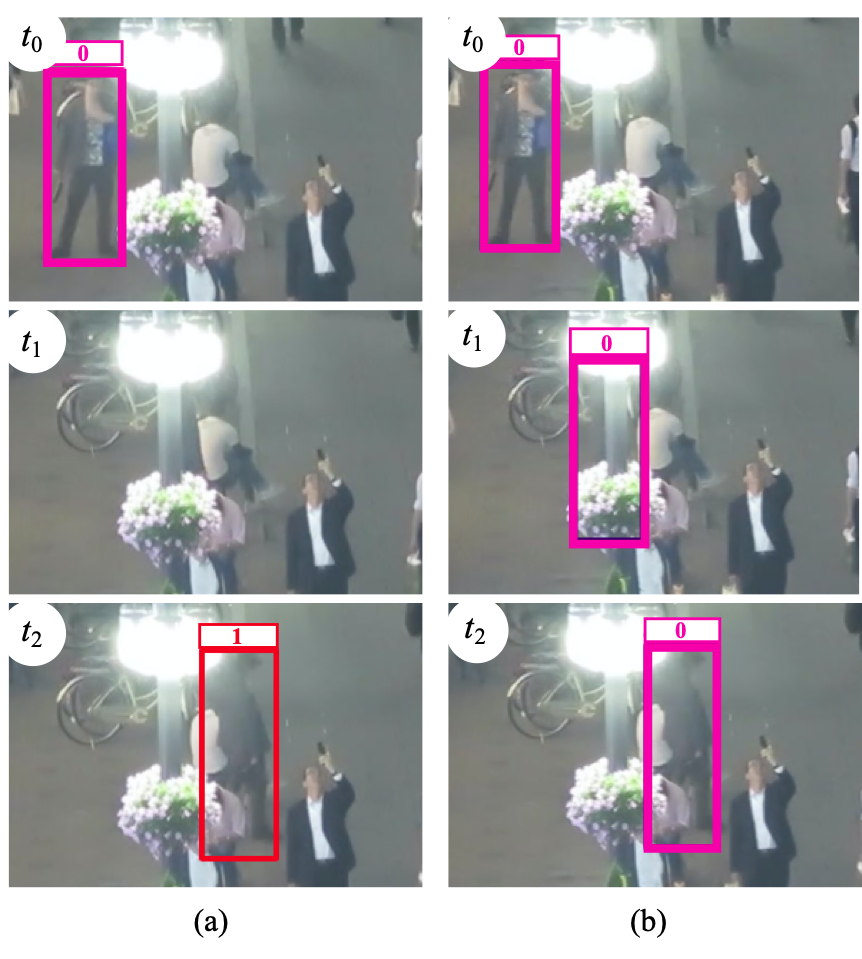}
\caption{A qualitative example showing (a) a failure case of MPNTrack\cite{braso2020learning} in handling long-term occlusions, which reduces the IDF1 score; (b) our method can effectively handle this case. The numbers are the object IDs. Best viewed in color.}
\label{fig:qualitative_example1}
\end{figure}

\begin{figure}[tp]
\centering
\includegraphics[width=0.42\textwidth]{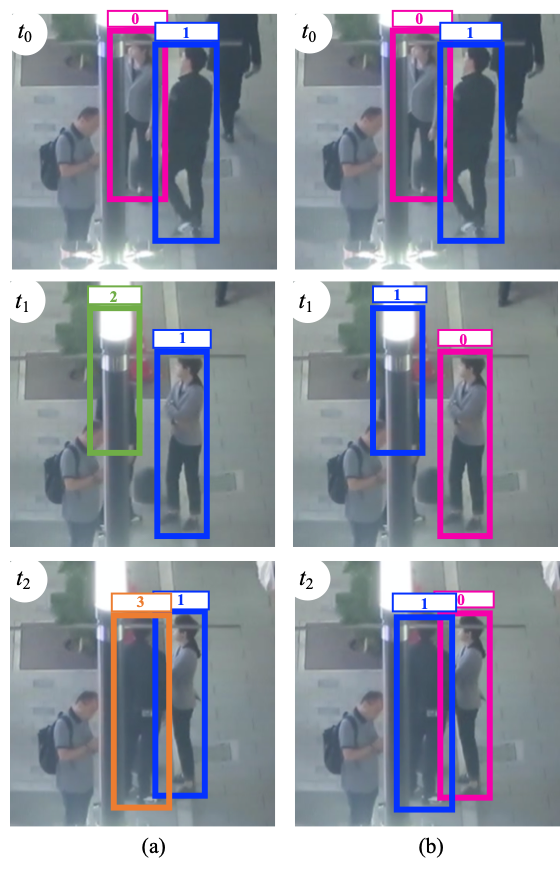}
\caption{A qualitative example showing (a) a failure case of MPNTrack~\cite{braso2020learning} in handling occlusions, which leads to an identity transfer when one person passes the other and a fragmentation when one is fully occluded; (b) our method can effectively handle this case. The numbers are the object IDs. Best viewed in color.}
\label{fig:qualitative_example2}
\end{figure}

Figure~\ref{fig:qualitative_example1} and Figure~\ref{fig:qualitative_example2} give a qualitative comparison between MPNTrack\cite{braso2020learning} and our method on MOT17. 
It validates that our method has better performance in handling long-term occlusions, hence achieving higher IDF1 score.

\section{Further Performance Comparison}~\label{section:supplementary_material_E}
We also noticed that MPNTrack~\cite{braso2020learning} used a different ReIdentification (ReID) model from our method.
In order to achieve a completely fair comparison, we also provide the comparison results between our method and MPNTrack using our ReID model on the training set of MOT17. Table~\ref{tab:mot17_train_eval} shows the detailed results. 
By comparing our method with MPNTrack$^{2}$, it is clear that our method achieves better performance on identity preservation, improving the IDF1 score by 1.5 percentage.
By comparing MPNTrack$^{1}$ with MPNTrack$^{2}$, we can see that the overall performance gap is small.
In summary, our method can achieve better association accuracy than MPNTrack~\cite{braso2020learning}.

\begin{table}[tbp]
\footnotesize
\centering
\begin{threeparttable}
 \setlength{\tabcolsep}{1.2mm}{
 \begin{tabular}{lcccccccc} 
  \toprule 
  Method  & MOTA$\uparrow$ & IDF1$\uparrow$ & MT$\uparrow$ & ML$\downarrow$ & FP$\downarrow$ & FN$\downarrow$ & IDs$\downarrow$  \\ 
  \midrule
   Ours & 63.9 & \textbf{71.8} & 647 & 377 & 7176 & \textbf{113700} & 728    \\
   MPNTrack\tnote{1}  & \textbf{64.0} & 70.0 & \textbf{648} & \textbf{362} & \textbf{6169} & 114509 & 602   \\
   MPNTrack\tnote{2} & 63.9 & 70.3 & 634 & 365 & 6228 & 114723 & \textbf{523}  \\
  \bottomrule 
 \end{tabular}}
 \begin{tablenotes}
     \item[1] with their own ReID model
     \item[2] with our ReID model
  \end{tablenotes}
 \vspace{10pt}
  \caption{\label{tab:mot17_train_eval}Further performance comparison on the training set of MOT17.} 
  \end{threeparttable}
\end{table}

\end{document}